\documentclass[11pt,a4paper]{article}
\usepackage[hyperref]{emnlp2020}
\usepackage{times}
\usepackage{latexsym}
\usepackage{amsmath}
\usepackage{amssymb}
\usepackage{float}
\usepackage{fancybox}
\usepackage{color}
\usepackage{enumitem}
\usepackage{multirow}
\usepackage[mathcal]{eucal}
\usepackage{tikz}
\usepackage{pgfplots}
\usetikzlibrary{shapes, arrows, positioning, decorations.markings}

\definecolor {processblue}{cmyk}{0.96,0,0,0}
\tikzstyle{int}=[draw, fill=blue!20, minimum size=2em]
\tikzstyle{init} = [pin edge={to-,thin,black}]

\usetikzlibrary{shapes}
\usetikzlibrary{fit}
\usetikzlibrary{chains}
\usetikzlibrary{arrows}

\tikzstyle{plate} = [draw, rectangle, rounded corners, fit=#1]
\tikzstyle{wrap} = [inner sep=0pt, fit=#1]

\tikzstyle{caption} = [node distance=0] %
\tikzstyle{bottom plate caption} = [caption, node distance=0, inner sep=0pt,
below left=-5pt and 0pt of #1.south east] %

\tikzstyle{top plate caption} = [caption, node distance=0, inner sep=0pt,
below left=0pt and 0pt of #1.north east] %

\usepackage{makecell}

\usepackage{microtype}

\aclfinalcopy 

\title{Neural Dialogue State Tracking with Temporally Expressive Networks}

\author{Junfan Chen \\
  BDBC and SKLSDE\\
  Beihang University, China \\
  \texttt{chenjf@act.buaa.edu.cn} \\
  \And
  Richong Zhang\thanks{$\ \ $Corresponding author}\\
  BDBC and SKLSDE\\
  Beihang University, China \\
  \texttt{zhangrc@act.buaa.edu.cn} \\
  \AND
Yongyi Mao \\
School of EECS\\
  University of Ottawa, Canada \\
  \texttt{ymao@uottawa.ca} \\
  \And 
  Jie Xu \\
  School of Computing\\
  University of Leeds, United Kingdom \\
  \texttt{j.xu@leeds.ac.uk} \\
}  

\date{}

\begin{document}
\maketitle

\begin{abstract}
Dialogue state tracking (DST) is an important part of a spoken dialogue system. Existing DST models either ignore temporal feature dependencies across dialogue turns or fail to explicitly model temporal state dependencies in a dialogue. In this work, we propose Temporally Expressive Networks (TEN) to jointly model the two types of temporal dependencies in DST. The TEN model utilizes the power of recurrent networks and probabilistic graphical models. Evaluating on standard datasets, TEN is demonstrated to improve the accuracy of turn-level-state prediction and the state aggregation. 

\end{abstract}
\section{Introduction}
Spoken dialogue systems (SDS) connect users and computer applications through human-machine conversations.  The users can achieve their goals, such as finding a restaurant, by interacting with a task-oriented SDS over multiple dialogue rounds or {\em turns}. 
Dialogue state tracking (DST) is an important task in SDS and the key function  is to maintain the {\em state} of the system so as to track the progress of the dialogue. 
In the context of this work, a state (or aggregated state) is the user's intention or interest accumulated from the conversation history, and the user's intention or interest at each turn is referred to as turn-level state. 

\begin{figure*}[ht]
	\begin{center}
		\begin{tabular}{c}
			\scalebox{0.9}{
				\begin{tikzpicture}[-latex ,auto ,node distance =1.2 cm and 2 cm, on grid , semithick ,
    	state/.style ={ circle , draw=red, fill=red!10, text=red , minimum width =1cm},
    	blank/.style ={ rectangle , color =white ,
    		draw=white!0 , text=black , minimum width =1cm, minimum height = 0.5cm},	
    	block/.style ={ rectangle , draw=white , text=black,
    		minimum width =0.2cm, minimum height = 1cm},	    				
    	box/.style ={rectangle , draw=blue, fill=blue!30 ,
    		draw, text=blue , minimum width =0.5cm , minimum height = 0.5cm},
    	cell/.style = {rectangle , draw=black, fill=blue!10 ,
    		 text=black , minimum width = 1cm , minimum height = 0.4cm}]

    	\foreach \r/\i in {0/1, 1.7/2, 3.4/3}
	    {
        	\node (b00) at (\r,0){};	
    		\node[cell](c00)[above=of b00, yshift=-0.3cm]{\bf FE};
    		\node[cell, fill=red!10](c01)[above=of c00]{\bf RC};
    		\node[cell, fill=green!10](c02)[above=of c01]{\bf CL};  
    		\node (b01) [above=of c02, yshift=-0.3cm]{};
    		
			\path (b00) edge node[left]{\small$a_{\i}$} node[right]{\small$u_{\i}$} (c00);   
			\path (c00) edge node[left]{\small$z_{\i}$} (c01); 	
			\path (c01) edge node[left]{\small$h_{\i}$} (c02); 		
			\path (c02) edge node[left]{\small$x_{\i}$} (b01);
			
		} 
		\draw[dotted, thick] (-1.2, 2.1) -- node[above]{\small$h_{0}$} (-0.5, 2.1);
		\draw[dotted, thick] (0.5, 2.1) -- node[above]{\small$h_{1}$} (1.2, 2.1);
		\draw[dotted, thick] (2.2, 2.1) -- node[above]{\small$h_{2}$} (2.9, 2.1);
		\node  (P0) at (0, 0){};
    	\node  (P0) at (0, 0){};	
		\node (b00) [above=of P0, xshift=1.66cm, yshift=-1.5cm]{(a) Implicit Tracking};  
    			
    	\foreach \r/\i in {5.6/1, 7.3/2, 9.0/3}
    	{
    		\node (b00) at (\r,0){};	
    		\node[cell](c00)[above=of b00, yshift=-0.3cm]{\bf FE};
    		\node[cell, fill=green!10](c01)[above=of c00]{\bf CL};
    		\node[cell, fill=yellow!10](c02)[above=of c01]{\bf SA};  
    		\node (b01) [above=of c02, yshift=-0.3cm]{};
    		
    		\path (b00) edge node[left]{\small$a_{\i}$} node[right]{\small$u_{\i}$} (c00);   
    		\path (c00) edge node[left]{\small$z_{\i}$} (c01); 	
    		\path (c01) edge node[left]{\small$y_{\i}$} (c02); 		
    		\path (c02) edge node[left]{\small$x_{\i}$} (b01);
    		
    	} 
    	\node  (P0) at (8, 0){};
		\draw[dashed, thick] (4.4, 3.3) -- node[above]{\small$x_{0}$} (5.1, 3.3);    	
    	\draw[dashed, thick] (6.1, 3.3) -- node[above]{\small$x_{1}$} (6.8, 3.3);
    	\draw[dashed, thick] (7.8, 3.3) -- node[above]{\small$x_{2}$} (8.5, 3.3);
	
    	\node  (P0) at (0, 0){};	
    	\node (b00) [above=of P0, xshift=7.32cm, yshift=-1.5cm]{(b) Explicit Tracking};
		
        \foreach \r/\i in {11.2/1, 12.9/2, 14.6/3}
        {
        	\node (b00) at (\r,0){};	
        	\node[cell](c00)[above=of b00, yshift=-0.3cm]{\bf FE};
        	\node[cell, fill=red!10](c01)[above=of c00]{\bf RC};
        	\node[cell, fill=green!10](c02)[above=of c01]{\bf CL};  
        	\node[cell, fill=yellow!10](c03)[above=of c02]{\bf SA};
        	\node(b01)[above=of c03, yshift=-0.3cm]{};
        	
        	\path (b00) edge node[left]{\small$a_{\i}$} node[right]{\small$u_{\i}$} (c00);   
        	\path (c00) edge node[left]{\small$z_{\i}$} (c01); 	
        	\path (c01) edge node[left]{\small$h_{\i}$} (c02); 		
        	\path (c02) edge node[left]{\small$p(y_{\i})$} (c03);
        	\path (c03) edge node[left]{\small$x_{\i}$} (b01);
        	
        } 
        \draw[dotted, thick] (10.0, 2.1) -- node[above]{\small$h_{0}$} (10.7, 2.1);
        \draw[dotted, thick] (11.7, 2.1) -- node[above]{\small$h_{1}$} (12.4, 2.1);
        \draw[dotted, thick] (13.4, 2.1) -- node[above]{\small$h_{2}$} (14.1, 2.1);
        \draw[dashed, thick] (10.0, 4.5) -- node[above]{\small$p(x_{0})$} (10.7, 4.5);
        \draw[dashed, thick] (11.7, 4.5) -- node[above]{\small$p(x_{1})$} (12.4, 4.5);
        \draw[dashed, thick] (13.4, 4.5) -- node[above]{\small$p(x_{2})$} (14.1, 4.5);
    	\node  (P0) at (0, 0){};	
    	\node (b00) [above=of P0, xshift=12.80cm, yshift=-1.5cm]{(c) Joint model (TEN)};        
        
		
\end{tikzpicture}
			}
		\end{tabular}
	\end{center}
	\caption{The model structures of Implicit Tracking, Explicit Tracking and Joint model. $(a, u)$:the system action and user utterance. $z$: features extracted from the $(a, u)$ pair. $h$:the hidden state of RNNs. $y$: the turn-level state. $x$: the aggregated state.   {\bf FE}:Feature Extractor, such as CNNs, RNNs. {\bf RC}:Recurrent Cell, such as LSTM, GRU. {\bf CL}:Classifier. {\bf SA}:State Aggregator. The dotted arrowed lines emphasize modeling temporal feature dependencies. The dashed arrowed lines emphasize modeling temporal state dependencies.}
	\label{fig:mod}
\end{figure*}
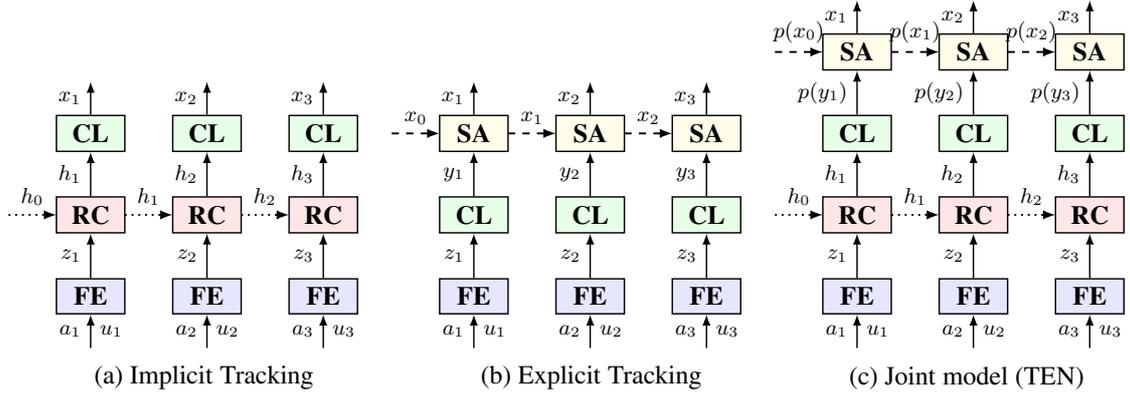

Many neural-network models have been successfully applied to DST. These models usually solve the DST problem by two approaches, the Implicit Tracking and the Explicit Tracking. As is shown in Figure~\ref{fig:mod} (a), the Implicit Tracking models ~\cite{Henderson:14b,Henderson:14,Mrksic:15,Ren:18,Ramadan:18,Lee:19} employs recurrent networks to accumulate features extracted from historical system action and user utterance pairs. A classifier is then built upon these accumulated features for state prediction. Although the Implicit Tracking captures temporal feature dependencies in recurrent-network cells, the state dependencies are not explicitly modeled. Only considering temporal feature dependencies is insufficient for accurate state prediction. This fact has been confirmed via an ablation study in our experiment.

Unlike the Implicit Tracking, the Explicit Tracking approaches, such as NBT~\cite{Mrksic:17} and GLAD~\cite{Zhong:18},  model the state dependencies explicitly. From the model structure in Figure~\ref{fig:mod}(b), the Explicit Tracking approaches first build a classifier to predict the turn-level state of each turn and then utilize a state aggregator for state aggregation. 

Despite achieving remarkable improvements upon the previous models, current Explicit Tracking models can be further improved in two aspects. One is that the temporal feature dependencies should be considered in model design. The Explicit Tracking models only extract features from the current system action and user utterance pair. In practice, the slot-value pairs in different turns are highly dependent. For example, if a user specifies $({\tt FOOD}, {\tt italian})$ at the current turn, he or she will probably not express it again in the future turns. For that reason, only extracting features from the current system action and user utterance pair is inadequate for turn-level state prediction. 

The other is that the uncertainties in the state aggregation can be more expressively modeled. The state-aggregation approaches in current Explicit Tracking models are sub-optimal. The deterministic rule in GLAD will propagate errors to future turns and lead to incorrect state aggregation. The heuristic aggregation in NBT needs further estimate the best configuration of its coefficient . An approach that can both reduce the error propagation and require less parameter estimation is necessary for the state aggregation.

In this study, we propose a novel Temporally Expressive Networks (TEN) to jointly model the temporal feature dependencies and temporal state dependencies (Figure~\ref{fig:mod} (c)). Specifically, to improve the turn-level state prediction, we exploits hierarchical recurrent networks to capture temporal feature dependencies across dialogue turns. Furthermore, to reduce state aggregation errors, we introduce factor graphs to formulate the state dependencies, and employ belief propagation to handle the uncertainties in state aggregation. Evaluating on the DSTC2, WOZ and MultiWoZ datasets, TEN is shown to improve the accuracy of the turn-level state prediction and the state aggregation. 
The TEN model establishes itself as a new state-of-the-art model on the DSTC2 dataset and a state-of-the-art comparable model on the WOZ dataset.

\section{Problem Statement}
In a dialogue system, the state is represented as a set of {\em slot-value} pairs. Let $\mathcal{S}$ denote the predefined set of slots.
For each slot $s\in \mathcal{S}$, let $\mathcal {V}(s)$  denote the set of all possible values associated with slot $s$. We also include an additional token, ${\tt unknown}$, as a legal value for all slots to represent their value is not determined. And we define 
 \begin{eqnarray*}
 \mathcal{V}^*(s) &:= &\mathcal {V}(s)\cup \{\tt unknown\}\\
 \mathcal{V}^* &:=&\bigcup_{s\in \mathcal {S}} \mathcal {V}^*(s) 
 \end{eqnarray*}
Let $\mathcal{X}$ denote the state space, and $x \in \mathcal{X}$ be a state configuration. Each state configuration $x$ can be regarded as a function mapping $x(s)$ from $\mathcal {S}$ to $\mathcal{V}^*$. For example,
\begin{equation}
 \label{eq:exampleState}
 x(s)= \left\{
 \begin{array}{ll}
    {\tt italian},  &s={\tt FOOD}  \\
    {\tt moderate},  & s={\tt PRICERANGE} \\
     {\tt unknown}, & s={\tt AREA}
 \end{array}
 \right.
\end{equation}
Let $x_t$ denotes the state configuration of the $t^{\rm th}$ dialogue turn, $u_t$ denotes the user utterance of the $t^{\rm th}$ turn and $a_t$ denotes the system action based on previous state $x_{t-1}$. Let $y_t \in \mathcal{X}$ be the turn-level state, which is meant to capture the user intention of the current utterance. The system computes the aggregated state $x_{t}$ through a deterministic procedure, according to $y_t$ and $x_{t-1}$. We next describe this procedure.

For any given $s$, we define an operator $\triangleleft$ on $\mathcal {V}^*(s)$ as follows. For any $v, v' \in \mathcal{V}^*(s)$,
\begin{equation}
v \triangleleft v' :=
\left\{
\begin{array}{cc}
     v, & {\rm if} ~v'={\tt unknown} \\
     v', & {\rm otherwise}
\end{array}
\right.
\end{equation}
We then extend the operator $\triangleleft$ to any two elements $x, y\in \mathcal{X}$, where $x\triangleleft y$ is also an element in $\mathcal{X}$ 
\begin{equation}
(x\triangleleft y)(s):=x(s)\triangleleft y(s).
\end{equation}
Using this notation, the aggregation of states is precisely according to 
\begin{equation}
    \label{eq:stateUpdate}
    x_t=x_{t-1} \triangleleft y_t.
\end{equation}
For example, if $x_{t-1}$ takes the configuration $x$ in (\ref{eq:exampleState}) and if $y_t$ is
\begin{equation}
 y_t(s)= \left\{
 \begin{array}{ll}
    {\tt chinese},  &s={\tt FOOD}  \\
    {\tt unknown},  & s={\tt PRICERANGE} \\
     {\tt unknown}, & s={\tt AREA}
 \end{array}
 \right.
\end{equation}
The aggregated state $x_t$ is
\begin{equation}
 x_{t}(s)= \left\{
 \begin{array}{ll}
    {\tt chinese},  &s={\tt FOOD}  \\
    {\tt moderate},  & s={\tt PRICERANGE} \\
     {\tt unknown}, & s={\tt AREA}
 \end{array}
 \right.
\end{equation}

The dialogue process can be characterized by a random process $\{(X_t, Y_t, A_t, U_t):t=1, 2, \ldots)\}$. In the DST problem, the probability measure ${\mathbb P}$ which defines the dialogue process is unknown.  We are however given a set $\mathcal{R}$ of realizations drawn from ${\mathbb P}$, where each $r\in \mathcal {R}$ is a dialogue, given in the form of $\{(x^{(r)}_t, y^{(r)}_t, a^{(r)}_t, u^{(r)}_t):t=1, 2, \ldots)\}$. Let  $x_{<t}$ denotes $(x_1, x_2, \ldots, x_t)$ and assume similar notations for $y_{<t}$, $a_{<t}$ etc. The learning problem for DST then becomes estimating ${\mathbb P}\left(x_t|a_{<t}, {u}_{<t}\right)$ for every $t$.



\section{Model}
This section introduces the proposed TEN model, which consists of Action-Utterance Encoder, Hierarchical Encoder, Turn-level State Predictor and State Aggregator. 
\begin{figure}[ht]
	\begin{center}
		\begin{tabular}{c}
			\scalebox{0.5}{
			\begin{tikzpicture}[-latex ,auto ,node distance =1.9 cm and 3.0 cm, on grid ,
    	semithick ,
    	oldstate/.style ={ circle ,top color =white , bottom color = red!20 ,
    		draw, red , text=red , minimum width =1.15cm},				
    	state/.style ={ circle , draw=red, fill=red!10, text=red , minimum width =1cm},
    	blank/.style ={ circle , color =white ,
    		draw, white , text=black , minimum width =1cm},				
    	box/.style ={rectangle ,top color =white , bottom color = processblue!20 ,
    		draw, processblue , text=blue , minimum width =0.1cm , minimum height = 0.1cm}]
				
		\node[state](X0){$X_{0}$};
		\node[state](X1)[right= of X0]{$X_{1}$};
		\node[state](Y1)[below= of X1]{$Y_{1}$};
		\node[state](H1)[below= of Y1]{$H_{1}$};
		\node[state](H0)[left= of H1]{$H_{0}$};
		\node[state](Z1)[below= of H1]{$Z_{1}$};
		\node[state](A1)[below= of Z1, xshift=-0.85cm]{$A_{1}$};
		\node[state](U1)[below= of Z1, xshift=0.85cm]{${U}_{1}$};
		
		\node[state](X2)[right= of X1]{$X_{2}$};
		\node[state](Y2)[below= of X2]{$Y_{2}$};
		\node[state](H2)[below= of Y2]{$H_{2}$};
		\node[state](Z2)[below= of H2]{$Z_{2}$};
		\node[state](A2)[below= of Z2, xshift=-0.85cm]{$A_{2}$};
		\node[state](U2)[below= of Z2, xshift=0.85cm]{${U}_{2}$};			
		
		\node[state](X3)[right= of X2]{$X_{3}$};
		\node[state](Y3)[below= of X3]{$Y_{3}$};
		\node[state](H3)[below= of Y3]{$H_{3}$};
		\node[state](Z3)[below= of H3]{$Z_{3}$};
		\node[state](A3)[below= of Z3, xshift=-0.85cm]{$A_{3}$};
		\node[state](U3)[below= of Z3, xshift=0.85cm]{${U}_{3}$};	
					
		\node[blank](B1)[right= of X3]{\Huge$\cdots$};	
		\node[blank](B2)[below= of B1]{\Huge$\cdots$};	
		\node[blank](B3)[below= of B2]{\Huge$\cdots$};
		\node[blank](B4)[below= of B3]{\Huge$\cdots$};
		\node[blank](B5)[below= of B4]{\Huge$\cdots$};	
													
		\path (X0) edge (X1);
		\path (X1) edge (X2);
		\path (Y1) edge (X1);
		\path (H0) edge (H1);		
		\path (H1) edge (Y1);
		\path (H1) edge (H2);
		\path (Z1) edge (H1);
		\path (A1) edge (Z1);
		\path (U1) edge (Z1);
		
		\path (X2) edge (X3);
		\path (Y2) edge (X2);	
		\path (H2) edge (Y2);
		\path (H2) edge (H3);
		\path (Z2) edge (H2);
		\path (A2) edge (Z2);
		\path (U2) edge (Z2);

		\path (X3) edge (B1);
		\path (Y3) edge (X3);	
		\path (H3) edge (Y3);
		\path (H3) edge (B3);
		\path (Z3) edge (H3);
		\path (A3) edge (Z3);
		\path (U3) edge (Z3);
				
\end{tikzpicture}
			}
		\end{tabular}
	\end{center}
	\caption{\label{fig:pgm} The probabilistic graphical model of TEN.}
\end{figure}
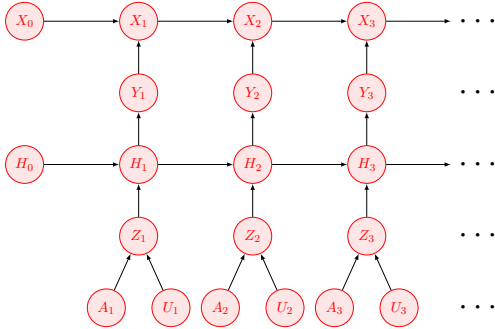
\subsection{Model Structure}
The overall model structure of TEN is shown in Figure~\ref{fig:mod} (c). we wish to express ${\mathbb P}(x_t| a_{<t}, {u}_{<t})$ using a probabilistic graphical model. For that purpose, we introduce two latent layers of random variables $\{H_t\}$ and $\{Z_t\}$, together with $\{Y_t\}$ and $\{X_t\}$, to form a Markov chain 
\begin{equation}
\label{eq:markov}
\{(A_t, {U}_t)\}\! \rightarrow \! \{Z_t\} \!\rightarrow \! \{H_t\} \!\rightarrow \!\{Y_t\}\! \rightarrow\! \{X_t\}.
\end{equation}
Then we can express the TEN model as a probabilistic graphical model shown in Figure~\ref{fig:pgm}. In the probabilistic graphical model, the variable $Z_t$ is a matrix of size $K_{\rm Z} \times |\mathcal{S}|$, each column of $Z_t(s)$ corresponds to a slot $s\in \mathcal{S}$. Obtained from $(A_t, {U}_t)$, $Z_t$ is referred to as the ``action-utterance encoding'' at turn $t$ which has a dimension of $K_{\rm Z}$. 
The variable $H_t$ is a matrix of size $K_{\rm H} \times |\mathcal{S}|$, with each column $H_t(s)$ also corresponding to the slot $s\in\mathcal{S}$. Here the recurrent $\{H_t\}$ layer is used to capture temporal feature dependencies, 
and $H_t$ is referred to as the "hierarchical encoding", which has a dimension of $K_{\rm H}$. In state aggregation, we introduce the factor graphs to model the state dependencies. The belief propagation is then employed to alleviate the error propagation. It allows the soft-label of $Y_t$ and $X_t$ keeping modeled. We next explain each module in detail. 
\subsubsection{Action-Utterance Encoder}
This module's function is to summarize the input system action and user utterance to a unified representation. For later use, we first define a GRU-attention encoder or abbreviated as GAE. The GAE block first feeds an arbitrary-length sequence of word-embedding vectors $(\overline{w}_1, \overline{w}_2, ..., \overline{w}_n):=\overline{w}_{<n}$ to a GRU encoder and obtains a hidden state vector $d_{i}$ at the $i^{th}$ time step, then weighted-combine all the hidden-state vectors using attention mechanism to construct the output vector $o$. The computation process of the GAE block is
\begin{equation}
\label{eq:gae}
\begin{split}
d_i&={\rm GRU}\left(d_{i-1}, \overline{w}_i; {\bf W}\right)\\
o &= \sum_{i=1}^n \frac{
	\exp \left(d_{i}^T \cdot \theta\right)
}
{\sum_{j=1}^n\exp \left(d_{j}^T \cdot \theta\right)
}d_{i}\\
\end{split}
\end{equation}
Here ${\bf W}$ is the parameter of the GRU networks and $\theta$ is the learnable parameter of attention mechanism. 
We simply introduce a notation ${\rm GAE}(\overline{w}_{<n}; {\bf W}, \theta)$ to indicate the above computation process (\ref{eq:gae}) of the GAE block. 

\noindent {\bf Utterance Encoder}. 
Let $\overline{w}^{{\rm u},t}_{<n}$ denotes the word-embedding sequence of the $t^{th}$ user utterance ${u}_t$. A GAE block is then used to obtain the utterance encoder with input $\overline{w}^{{\rm u},t}_{<n}$.  
For each slot $s\in \mathcal{S}$, an utterance encoding $\overline{u}_t(s)$ is computed by
\begin{equation}
\label{eq:ue}
\overline{u}_t(s)={\rm GAE}(\overline{w}^{{\rm u}, t}_{<n}; {\bf W}_{\rm u}, \theta_s)
\end{equation}
Note that the GAEs for different slot $s$ share the same parameter ${\bf W}_{\rm u}$, but they each have their own attention parameter $\theta_s$.  

\noindent {\bf Action Encoder}. The system action at each turn may contain several phrases~\cite{Zhong:18}. Suppose that action $a_t$ contains $m$ phrases. Each phrase $b_t^i \in a_t$ is then taken as a word sequence, and let its word-embedding sequence be denoted as $\overline{b}_t^i$. For each $i$ and each slot
$s$,  $\overline{b}_t^i$
is passed to a GAE block and the action-phrase vector $c_t^i(s)$ is computed by 
\begin{equation}
\label{eq:ae}
c_t^i(s)={\rm GAE}(\overline{b}_t^i; {\bf W}_{\rm a}, \varphi_s)
\end{equation}
Like utterance encoder, these $|\mathcal{S}|$ parallel GAE's share the same GRU parameter ${\bf W}_{\rm a}$ but each has its own attention parameters $\varphi_s$. Finally, we adopt the same approach proposed in ~\cite{Zhong:18}, which combines the action-phrase vectors to a single vector by attention mechanism. Specifically, the action encoding $\overline{a}_t(s)$ is obtained by interacting with utterance encoding $\overline{u}_t(s)$, calculated as
\begin{equation}
\overline{a}_t(s) = \sum_{i=1}^m \frac{
\exp \left(\overline{u}_t(s)^T \cdot c_t^i(s)\right)
}
{\sum_{j=1}^m\exp \left(\overline{u}_t(s)^T \cdot c_t^j(s)\right)
}c_t^i(s)
\end{equation}
\noindent {\bf Action-utterance Encoding}. The action-utterance encoding $z_t(s)$ is simply the concatenation of vectors $\overline{u}_t(s)$ and $\overline{a}_t(s)$. 

\subsubsection{Hierarchical Encoder}
Instead of only utilizing the current action-utterance encoding for turn-level state prediction, in this module, we introduce the hierarchical recurrent networks to model the temporal feature dependencies across turns. Specifically, upon the GAE blocks, we use $|\mathcal{S}|$ parallel GRU networks to obtain the hierarchical encoding $\{h_t\}$ from all the historical action-utterance encoding vectors. The hierarchical encoding for each slot $s$ is computed by
\begin{equation}
h_t(s)={\rm GRU}\left(h_{t-1}(s), z_{t}(s); {\bf W}_{\rm h}\right)
\end{equation}
where the parameter ${\bf W}_{\rm h}$ of these GRU networks, is shared across all slots. 
\subsubsection{Turn-level State Predictor}
The Turn-level State Predictor is simply implemented by $|\mathcal{S}|$ softmax-classifiers, each for a slot $s$ according to 
\begin{equation}
\label{eq:h2y}
{\mathbb P}\left(y_t(s)|a_{<t}(s), 
{u}_{<t}(s)
\right)
:={\rm smax}\left(\phi^T_s h_t(s) \right)
\end{equation}
where ${\rm smax}$ denote the softmax function and $\phi_s$ with size $K_{\rm h} \times |\mathcal{V}^{*}(s)|$ serves as the weight matrix of the classifiers. We will denote this predictive distribution for turn-level state $y_t(s)$ computed by (\ref{eq:h2y}) as $\alpha_t^s$.
\subsubsection{State Aggregator}
One of the insights in this work is that when a hard decision is made on the soft-label, the errors it creates may propagate to future turns, resulting in errors in future state aggregation. We insist that the soft-label of $Y_t$ and $X_{t}$ should be maintained so that the uncertainties in state aggregation can be kept in modeling. Thus we propose a state aggregator based on the factor graphs and handle these uncertainties using belief propagation.

\noindent {\bf Factor Graphs}.
For utilizing the factor graphs in state aggregation, we first introduce an indicator function, denoted by $g$, according to the deterministic aggregation rule $\triangleleft$. Specifically, 
for any $v, v', v'' \in \mathcal {V}^*(s)$, 
\begin{equation}
    g(v, v', v''):= \left\{
    \begin{array}{ll}
    1, & {\rm if}~ v\triangleleft v'= v''\\
    0, &{\rm otherwise}
    \end{array}
    \right.
\end{equation}

\begin{figure}[ht]
	\begin{center}
		\begin{tabular}{c}
			\scalebox{0.6}{
				\begin{tikzpicture}[-latex ,auto ,node distance =2 cm and 2 cm, on grid ,
    	semithick ,
    	state/.style ={ circle , draw=black, fill=red!20, text=black , minimum width =1cm},
    	blank/.style ={ circle , color =white ,
    		draw, white , text=black , minimum width =1cm},				
    	box/.style ={rectangle , draw=black, fill=blue!20 ,
    		draw, text=black , minimum width =0.5cm , minimum height = 0.5cm}]
    	
    
    	\node[state](X0){$x_{0}(s)$};
    	\node[box](g1)[right=of X0]{};
    	\node[above=0.1cm] at (g1.north) {$g$};
    	\node[state](L1)[below=of g1]{$y_{1}(s)$}; 
    	\node[box](p1)[below=of L1]{};
    	\node[below=0.3cm] at (p1.south) {$\alpha_1^s$};
    	\node[state](X1)[right=of g1]{$x_{1}(s)$};
    	\node[box](g2)[right=of X1]{};
    	\node[above=0.1cm] at (g2.north) {$g$};
    	\node[state](L2)[below=of g2]{$y_{2}(s)$}; 
    	\node[box](p2)[below=of L2]{};
    	\node[below=0.3cm] at (p2.south) {$\alpha_2^s$};				
    	\node[state](X2)[right=of g2]{$x_{2}(s)$};	
    	
    
    	\node[blank](A)[right= of X2]{\Huge$\cdots$};
    	\node[blank](B)[below= of A]{\Huge$\cdots$};
       	\node[blank](C)[below= of B]{\Huge$\cdots$};
    	
    	
    	\path[-] (X0) edge (g1);
    	\path ([xshift=0.1cm,yshift=-0.2cm]X0.east) edge node[below]{\footnotesize$\gamma_{1}^s$} ([xshift=-0.1cm, yshift=-0.2cm]g1.west);		
    	\path[-] (g1) edge (L1);
    	\path ([xshift=0.2cm,yshift=0.1cm]L1.north) edge node[right]{\footnotesize$\beta_{1}^s$} ([xshift=0.2cm, yshift=-0.1cm]g1.south);			
    	\path[-] (L1) edge (p1);
    	\path ([xshift=0.2cm,yshift=0.1cm]p1.north) edge node[right]{\footnotesize$\alpha_{1}^s$} ([xshift=0.2cm, yshift=-0.1cm]L1.south);				
    	\path[-] (g1) edge (X1);
    	\path ([xshift=0.1cm,yshift=-0.2cm]g1.east) edge node[below]{\footnotesize$\mu_{1}^s$} ([xshift=-0.1cm, yshift=-0.2cm]X1.west);				
    	\path[-] (X1) edge (g2);
    	\path ([xshift=0.1cm,yshift=-0.2cm]X1.east) edge node[below]{\footnotesize$\gamma_{2}^s$} ([xshift=-0.1cm, yshift=-0.2cm]g2.west);					
    	\path[-] (g2) edge (L2);
    	\path ([xshift=0.2cm,yshift=0.1cm]L2.north) edge node[right]{\footnotesize$\beta_{2}^s$} ([xshift=0.2cm, yshift=-0.1cm]g2.south);												
    	\path[-] (L2) edge (p2);
    	\path ([xshift=0.2cm,yshift=0.1cm]p2.north) edge node[right]{\footnotesize$\alpha_{2}^s$} ([xshift=0.2cm, yshift=-0.1cm]L2.south);				
    	\path[-] (g2) edge (X2);
    	\path ([xshift=0.1cm,yshift=-0.2cm]g2.east) edge node[below]{\footnotesize$\mu_{2}^s$} ([xshift=-0.1cm, yshift=-0.2cm]X2.west);	
    	
\end{tikzpicture}
			}
		\end{tabular}
	\end{center}
	\caption{The factor graph for $G(x_{<t}(s), y_{<t}(s))$.}
	\label{fig:simplefactor}
\end{figure}
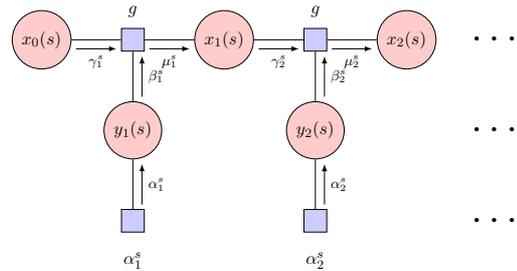

According to the probabilistic graphical model expressed in Figure~\ref{fig:pgm}, it can be derived that
{\small
\begin{eqnarray}
\nonumber
&\!\!\!\!{\mathbb P}&(x_t|a_{<t}, {u}_{<t}) \\
\nonumber
& \!\!\!\!\!\!\!\!= \!\!\! & \!\!\!\!\!
\sum\limits_{x_{<t-1}} \sum\limits_{y_{<t}} 
\prod\limits_{s\in \mathcal{S}} 
\alpha_t^s(y_t(s)) \prod\limits_{\tau=1}^t g(x_{\tau-1}(s), y_{\tau}(s), x_{\tau}(s)) \\
\nonumber
& \!\!\!\!\!\!\!\!= \!\!\! & \!\!\!\!\!
\prod\limits_{s\in \mathcal{S}}
\underbrace{
\sum\limits_{x_{<t-1}(s)} \sum\limits_{y_{<t}(s)} 
\underbrace{
 \!\!\!\!\alpha_t^s(y_t(s)) \!\!\prod\limits_{\tau=1}^t g(x_{\tau-1}(s), y_{\tau}(s), x_{\tau}(s))}_{G\left(x_{<t}(s), y_{<t}(s)\right)}
 }_{Q_t^s(x_t(s))} 
\end{eqnarray}
}
where the term $Q_t^s(x_t(s))$ above is precisely
${\mathbb P}(x_t(s)|a_{<t}, {u}_{<t})$,
a distribution on $\mathcal{V}^*(s)$. 
It turns out that the term $G(x_{<t}(s), y_{<t}(s))$ in the double summation of $Q_t^s(x_t(s))$, despite its complexity, can be expressed elegantly using a factor graph in Figure~\ref{fig:simplefactor}. 


\noindent {\bf Belief Propagation}.
Factor graphs are powered by a highly efficient algorithm, called the belief propagation or the sum-product algorithm, for computing the marginal distribution.
In particular, the algorithm executes by passing ``messages" along the edges of the factor graph 
and the sent message is computed from all incoming messages on its ``upstream".  
For a detailed description of message computation rules in belief propagation, the reader is referred to~\cite{Kschischang:01}. 

Applying the principle of belief propagation, one can also efficiently express $Q_t^s$ at each turn $t$ for each slot $s$ in terms of message passing. We now describe this precisely. 

Let $T$ denote the total number of turns of the dialogue. For each slot $s$, a factor graph representation $G(x_{<T}(s), y_{<T}(s))$ can be constructed. For each $t=1, \ldots, T$, let messages $\beta_t^s$, $\gamma_t^s$ and $\mu_t^s$ be introduced on the edges of the factor graph as shown in Figure~\ref{fig:simplefactor} and the computation of these messages are given below.
\begin{equation}
\label{eq:fg}
\!\!\!\!\!\left\{
\begin{array}{lll}
\beta_{t}^s &\!\!\!\!\!:= & \alpha_{t}^s\\
\gamma_{t}^s &\!\!\!\!\!:= &\mu_{t-1}^s\\
\mu_{t}^s (v) &\!\!\!\!\! := &\!\!\!\!\!\!\! \!\!\!\!\!\!\! \sum\limits_{(v', v'') \in 
\mathcal{V}^*(s) \times \mathcal{V}^*(s)
}\!\!\!\!\! \!\!\!\!\!\!\!
g(v', v'', v)\gamma^s_{t} (v')\beta^s_{t} (v'')
\end{array}
\right.
\end{equation}
where $\mu_0^s$ is defined by
\begin{equation*}
\mu_0^s(v) = 
\left\{
\begin{array}{ll}
1,& {\rm if} ~v={\tt unknown}\\ 
0,&{\rm otherwise}.
\end{array}\right.
\end{equation*}

According to message computation rule given in (\ref{eq:fg}), for each $t\le T$ and each slot $s\in  \mathcal{S}$, $\mu_t^s=Q_t^s.$
Recalling that $Q_t^s$ is the predictive distribution for state $x_t(s)$ and $\alpha_{t}^s$ is the predictive distribution for turn-level state $y_t(s)$, we have completed specifying how the factor graphs and the belief propagation are utilized for state aggregation.

\subsection{Loss Function and Training}

Under the TEN model, the cross-entropy loss on the training set $\mathcal{R}$ follows the standard definition as below
\begin{equation}
    \mathcal{L}_{\rm TEN}:= \sum\limits_{r\in {\mathcal {R}}} \sum\limits_{s\in \mathcal {S}}\sum\limits_{t=1}^{T^{(r)}} -\log Q_t^s(x_t^{(r)}(s))
\end{equation}
where the superscript ``$(r)$'' indexes a training dialogue in $\mathcal{R}$. 
It is worth noting that this loss function, involving the message computation rules, can be directly optimized by the stochastic gradient descent (SGD) method.

For ablation studies, we next present three ablated versions of the TEN model.

\noindent{\bf TEN--Y Model} In this model, we discard the $\{Y_t\}$ layer of TEN (hence the name TEN--Y) and conduct state aggregation using RNNs. The model then turns to be an Implicit Tracking model. The state distribution ${\mathbb P}(x_t(s)|a_{<t}, {u}_{<t})$ is computed directly by the softmax-classifiers in (\ref{eq:h2y}). We will denote the state distribution computed this way by $\widetilde{Q}_t^s$. The cross-entropy loss is then defined as
\begin{equation}
    \mathcal{L}_{\rm TEN-Y}:= \sum\limits_{r\in {\mathcal {R}}} \sum\limits_{s\in \mathcal {S}}\sum\limits_{t=1}^{T^{(r)}} -\log \widetilde{Q}_t^s(x_t^{(r)}(s))
\end{equation}

\noindent{\bf TEN--X Model} In this model, instead of training against the state sequence $\{x_t\}$, the training target is taken as 
the corresponding turn-level state sequence $\{y_t\}$. The computation of $\{x_t\}$ can be done through the operator $\triangleleft$ : $x_t = x_{t-1}\triangleleft y_t$.
When using the turn-level state as training target, one discards the $\{X_t\}$ layer of TEN (hence the name TEN--X). The difference between TEN--X and TEN is that TEN--X aggregate states using the deterministic rule $\triangleleft$ while TEN using the factor graphs. The cross-entropy loss for TEN--X is naturally defined as
\begin{equation}
    \mathcal{L}_{\rm TEN-X}:= \sum\limits_{r\in {\mathcal {R}}} \sum\limits_{s\in \mathcal {S}}\sum\limits_{t=1}^{T^{(r)}} 
    -\log \alpha_t^s(y_t^{(r)}(s))
\end{equation}

\noindent{\bf TEN--XH Model} In this model, the Hierarchical Encoder layer $\{H_t\}$ is removed from TEN--X, and the model is reduced to an Explicit Tracking mode. In this case, the computation of $\alpha_t^s$ (or ${\mathbb P}(y_t(s)|a_{<t}, \widetilde{u}_{<t})$) in (\ref{eq:h2y}) is done by replacing the input $h_t(s)$ with the action-utterance encoding $z_t(s)$.  We will denote the $\alpha_t^s$ computed this way by $\widetilde{\alpha}_t^s$. The TEN--XH and TEN--X models are different in whether the temporal feature dependencies are considered or not. The cross-entropy loss for TEN--XH is
\begin{equation}
    \mathcal{L}_{\rm TEN-XH}:= \sum\limits_{r\in {\mathcal {R}}} \sum\limits_{s\in \mathcal {S}}\sum\limits_{t=1}^{T^{(r)}} 
    -\log \widetilde{\alpha}_t^s(y_t^{(r)}(s))
\end{equation}

\section{Experiment}
\subsection{Datasets}
The second Dialogue State Tracking Challenge dataset (DSTC2)~\cite{Henderson:14a}, the second version of the Wizard-of-Oz dataset (WOZ)~\cite{Lina:17} and MultiDomain Wizard-of-Oz dataset (MultiWOZ)~\cite{Budzianowski:18} are used to evaluate the models. Both the DSTC2 and WOZ datasets contain conversations between users and task-oriented dialogue systems about finding suitable restaurants around Cambridge. The DSTC2 and WOZ datasets share the same ontology, which contain three informable slots: ${\tt FOOD}$, ${\tt AREA}$, ${\tt PRICERANGE}$. 
The official DSTC2 dataset contains some spelling errors in the user utterances, as is pointed out in~\cite{Mrksic:17}. Thus we use the manually corrected version provided by~\cite{Mrksic:17}. This dataset consists of $ 3,235 $ dialogues with $ 25,501 $ turns. There are $ 1,612 $ dialogues for training, $ 506 $ dialogues for validation and $ 1,117 $ dialogues for testing. The average turns per dialogue is $ 14.49 $.
In the WOZ dataset, there are $ 1,200 $ dialogues with $ 5,012 $ turns. The number of dialogues used for training, validation and testing are $600$, $200$ and $400$ respectively. The average turns per dialogue is $ 4 $.
The MultiWOZ dataset is a large multi-domain dialogue state tracking dataset with $30$ slots, collected from human-human conversations. The training set contains $8,438$ dialogues with $115,424$ turns. There are respectively $1,000$ dialogues in validation and test set. The average turns per dialogue is $ 13.68$.
\begin{table}[ht]
	\centering
	\caption{Joint goal accuracy on the DSTC2, WOZ and MultiWOZ dataset.}
	\setlength{\tabcolsep}{3pt}
	\begin{tabular}{|c|c|c|c|}
		\hline
		\bf{Model} & DSTC2 & WOZ & MultiWOZ\\		
		\hline
		NBT-DNN & 72.6 & 84.4 & -\\
		\hline
		NBT-CNN & 73.4 & 84.2 & -\\		
		\hline
		Scalable & 70.3 & - & -\\		
		\hline	
		MemN2N & 74.0 & - & -\\
		\hline			
		PtrNet & 72.1 & - & -\\		
		\hline	
		LargeScale & - & 85.5 & 25.8\\				
		\hline				
		GLAD & 74.5 & 88.1 & 35.6\\
		\hline
		GCE & - & 88.5 & 35.6\\
		\hline
		StateNetPSI & 75.5 & 88.9 & -\\
		\hline
		SUMBT & - & \textbf{91.0} & 42.4\\		
		\hline
		HyST & - & - & 44.2\\		
		\hline
		DSTRead+JST & - & - & 47.3\\	
		\hline
		TRADE & - & - & 48.6 \\	
		\hline
		COMER & - & 88.6 & 45.7\\		
		\hline	
		DSTQA & - & - & \bf{51.4} \\		
		\hline	
		MERET & - & - & 50.9\\		
		\hline	
		SST & - & - & 51.2\\		
		\hline		
		\bf{TEN--XH}& 73.5 & 88.8 & 42.0\\		
		\hline
		\bf{TEN--Y}& 74.7 & 89.6 & 45.9\\
		\hline	
		\bf{TEN--X}& 76.2 & 89.3 & 46.3\\
		\hline		
		\bf{TEN} & \bf{77.3} & 90.8 & 46.6 \\		
		\hline										
	\end{tabular}
	\label{tab:res_joint_goal}
\end{table} 
\subsection{Evaluation Metrics and Compared Models}
In this work, we focus on the standard evaluation metrics, {\em joint goal accuracy}, which is described in~\cite{Henderson:14a}. The {\em joint goal accuracy} is the proportion
of dialogue turns whose states are correctly predicted. In addition, we also report the {\em turn-level state accuracy} of TEN--XH and TEN--X model for ablation studies. 

The models used for comparison include NBT-DNN~\cite{Mrksic:17}, NBT-CNN~\cite{Mrksic:17}, Scalable~\cite{Rastogi:17}, MemN2N~\cite{Liu:17a}, PtrNet~\cite{Xu:18}, LargeScale~\cite{Ramadan:18}, GLAD~\cite{Ramadan:18}, GCE~\cite{Nouri:18}, StateNetPSI~\cite{Ren:18}, SUMBT~\cite{Lee:19}, HyST~\cite{Goel:19}, DSTRead+JST~\cite{Gao:19}, TRADE~\cite{Wu:19}, COMER~\cite{Ren:19}, DSTQA~\cite{Zhou:19}, MERET~\cite{Huang:20} and SST~\cite{Chen:20}.
\subsection{Implementation}
The proposed models are implemented using the Pytorch framework. The code and data are released on the Github page\footnote{https://github.com/BDBC-KG-NLP/TEN\_EMNLP2020}. The word embedding is the concatenation of the pre-trained GloVe embeddings~\cite{Pennington:14} and the character n-gram embeddings~\cite{Hashimoto:17}. 
We tune the hyper-parameters by grid search on the validation set. 
The GAE block is implemented with bi-directional GRUs, and the hidden state dimension of the GAE is 50. The hidden state dimension of the GRU used in the Hierarchical Encoder module is 50. The fixed learning rate is 0.001. The Adam optimizer~\cite{Kingma:15} with the default setting is used to optimize the models.
It is worth mentioning that the TEN model can be difficult to train with SGD from a cold start. This is arguably due to the ``hard'' $g$ function. That is, the $\{0, 1\}$-valued nature of $g$ is expected to result in sharp barriers in the loss landscape,  preventing gradient-based optimization to cross.  Thus when training TEN, we start with the parameters obtained from a pre-trained TEN--X model.
\subsection{Evaluation Results}
The joint goal accuracy results on the DSTC2,WOZ and MultiWOZ datasets are shown in Table~\ref{tab:res_joint_goal}. From the table, we observe that the proposed TEN model outperforms previous models on both DSTC2 and WOZ datasets, except SUMBT, a model boosted with pre-trained BERT~\cite{Devlin:19} model. It is worth noting that TEN, built upon attention-based GRU encoders, achieves comparable performance with SUMBT, without incorporating pre-trained language models.  This fact demonstrates that TEN is a strong model for DST. 
Comparing to TEN--XH, the TEN--X model obtains impressive $2.7\%$, $0.5\%$ and $4.3\%$ performance gains on the DSTC2, WOZ and MultiWOZ dataset respectively. These performance gains demonstrate that the state estimation benefits from more accurate turn-level state prediction. The TEN model further improves upon the TEN--X model by $1.1\%$ on the DSTC2 dataset, $1.5\%$ on the WOZ dataset and $0.3\%$ on the MultiWOZ dataset. The TEN model achieves these improvements by modeling uncertainties with the belief propagation in the state aggregation. Although both TEN--Y and TEN have modeled the temporal feature dependencies, TEN--Y performs much worse than TEN. This fact indicates that only considering temporal feature dependencies is inadequate for DST. Models relying on pre-defined ontologies (including GLAD,GCE,SUMBT and TEN) suffer from computational complexity when applying to multi-domain DST datasets with a large set of slots~\cite{Ren:19}, which leads to worse performance than recent generation-based models (DSTRead+JST,TRADE,DSTQA,MERET and SST, specially designed for multi-domain DST) on the MultiWOZ dataset.
\begin{table*}[ht]
    \small
	\centering
	\caption{An example of dialogue state tracking. We only report the results from turn 1 to turn 4 on slot $s={\tt FOOD}$ and focus on ${\tt dontcare}$(${\tt dcr}$) and ${\tt unknown}$(${\tt unk}$) value due to space limitation. \textbf{S} and \textbf{U} represent the system utterance and the user utterance, respectively. The boldface emphasizes the highest-probability value.}
	\setlength{\tabcolsep}{3pt}
	\begin{tabular}{|c|c|c|c|c|c|c|c|}
		\hline
		$t$ & $(a_t, u_t)$ & $\alpha_t^{s}$ & $y_{t}(s)$ & $Q_t^{s}$& $x_{t}^{s}$ & TEN--X & TEN  \\		
		\hline
		1& \makecell*[l]{\textbf{S}:welcome to cambridge restaurant system.\\\textbf{U}:im looking for a moderately priced} & \makecell*[c]{$({\tt dcr}, 0.00)$\\$\bf{(unk, 0.99)}$} & ${\tt unk}$ & \makecell*[c]{$({\tt dcr}, 0.00)$\\$\bf{(unk, 0.99)}$} & ${\tt unk}$ & ${\tt unk}$ & ${\tt unk}$\\
		\hline
		2& \makecell*[l]{\textbf{S}:moderate price range. what type of food do you want?\\\textbf{U}:restaurant and it should be} & \makecell*[c]{$({\tt dcr}, 0.00)$\\$\bf{(unk, 0.48)}$} & ${\tt unk}$ & \makecell*[c]{$({\tt dcr}, 0.00)$\\$\bf{(unk, 0.48)}$} & ${\tt unk}$ & ${\tt unk}$ & ${\tt unk}$\\
		\hline	
		3& \makecell*[l]{\textbf{S}:you want a restaurant serving any type of food right?\\\textbf{U}:yea} & \makecell*[c]{$({\tt dcr}, 0.45)$\\$\bf{(unk, 0.54)}$} & ${\tt dcr}$ & \makecell*[c]{$\bf{(dcr, 0.45)}$\\$({\tt unk}, 0.26)$} & ${\tt dcr}$ & ${\tt unk}$ & ${\tt dcr}$\\
		\hline	
		4& \makecell*[l]{\textbf{S}:what part of town do you have in mind?\\\textbf{U}:north} & \makecell*[c]{$({\tt dcr}, 0.00)$\\$\bf{(unk, 0.99)}$} & ${\tt unk}$ & \makecell*[c]{$\bf{(dcr, 0.45)}$\\$({\tt unk}, 0.26)$} & ${\tt dcr}$ & ${\tt unk}$ & ${\tt dcr}$\\
		\hline								
	\end{tabular}
	\label{tab:ex_bp}
\end{table*}
\begin{figure}[ht]
	\begin{center}
		\begin{tabular}{c}
			\scalebox{0.8}{
				\begin{tikzpicture}
	\begin{axis}[
		xlabel=Turns,
		ylabel=Joint Goal,
		legend pos=south west
		]
	\addplot[color=black,mark=x] coordinates {
		(1, 0.7878)
		(2, 0.7377)
		(3, 0.7601)
		(4, 0.7769)
		(5, 0.7819)
		(6, 0.7784)
		(7, 0.7567)
		(8, 0.7327)
		(9, 0.7051)
		(10, 0.7016)
		(11, 0.6741)
		(12, 0.6262)
		(13, 0.5854)
		(14, 0.5794)
		(15, 0.5490)
		(16, 0.4706)
		(17, 0.4853)
		(18, 0.5172)
		(19, 0.4808)
		(20, 0.475)
		(21, 0.5152)
		(22, 0.4643)
		(23, 0.5455)
		(24, 0.5263)
		(25, 0.6)
		(26, 0.6667)
		(27, 0.5714)
	};
	\addplot[color=blue,mark=*] coordinates {
		(1, 0.7896)
		(2, 0.7386)
		(3, 0.7663)
		(4, 0.7913)
		(5, 0.8012)
		(6, 0.7996)
		(7, 0.7888)
		(8, 0.7817)
		(9, 0.7714)
		(10, 0.75)
		(11, 0.7111)
		(12, 0.6699)
		(13, 0.6280)
		(14, 0.6111)
		(15, 0.6176)
		(16, 0.5529)
		(17, 0.5735)
		(18, 0.6379)
		(19, 0.5769)
		(20, 0.625)
		(21, 0.7576)
		(22, 0.7143)
		(23, 0.7273)
		(24, 0.7368)
		(25, 0.7333)
		(26, 0.8333)
		(27, 0.7143)
	};	
	\addplot[color=green,mark=triangle*] coordinates {
		(1, 0.7816)
		(2, 0.7234)
		(3, 0.7395)
		(4, 0.7633)
		(5, 0.7695)
		(6, 0.7661)
		(7, 0.7674)
		(8, 0.7682)
		(9, 0.7564)
		(10, 0.7473)
		(11, 0.7259)
		(12, 0.7039)
		(13, 0.7073)
		(14, 0.6587)
		(15, 0.6569)
		(16, 0.6118)
		(17, 0.5735)
		(18, 0.6207)
		(19, 0.6346)
		(20, 0.7)
		(21, 0.7273)
		(22, 0.6429)
		(23, 0.6364)
		(24, 0.5789)
		(25, 0.6)
		(26, 0.8333)
		(27, 0.7143)
	};		
	\addplot[color=red,mark=square*] coordinates {
		(1, 0.7950)
		(2, 0.7520)
		(3, 0.7726)
		(4, 0.7986)
		(5, 0.8079)
		(6, 0.8107)
		(7, 0.8008)
		(8, 0.7902)
		(9, 0.7735)
		(10, 0.7661)
		(11, 0.7333)
		(12, 0.6942)
		(13, 0.6707)
		(14, 0.6508)
		(15, 0.6569)
		(16, 0.6118)
		(17, 0.6176)
		(18, 0.6207)
		(19, 0.5962)
		(20, 0.675)
		(21, 0.7879)
		(22, 0.6786)
		(23, 0.6818)
		(24, 0.6842)
		(25, 0.6667)
		(26, 0.75)
		(27, 0.7143)
	};		
	\legend{TEN--XH,TEN--X,TEN--Y,TEN}
	\end{axis}
\end{tikzpicture}
			}
		\end{tabular}
	\end{center}
	\caption{Temporal analysis on the DSTC2 dataset.}
	\label{fig:seq}
\end{figure}
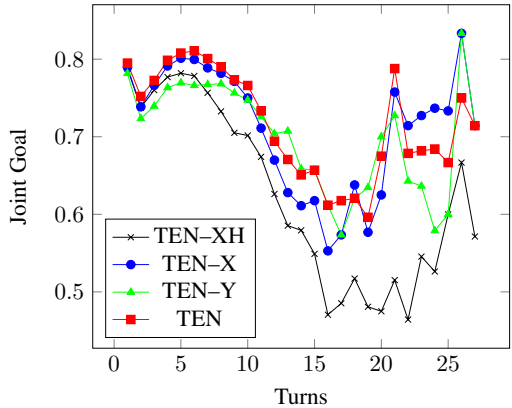
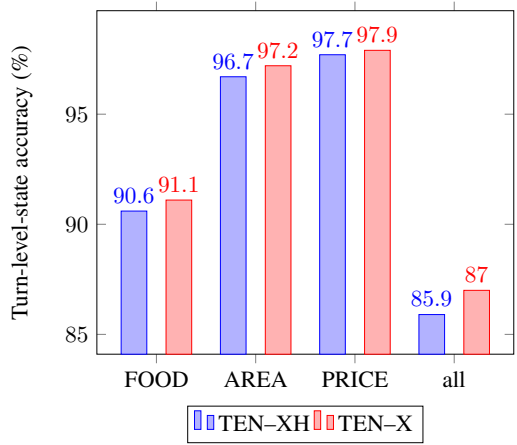
\begin{figure}[ht]
	\begin{center}
		\begin{tabular}{c}
			\scalebox{0.8}{
				\begin{tikzpicture}
\begin{axis}[
	x tick label style={
		/pgf/number format/1000 sep=},
	ylabel=Turn-level-state accuracy (\%),
	enlargelimits=0.15,
	enlarge x limits=0.2,
	legend style={at={(0.5,-0.15)},
		anchor=north,legend columns=-1},
	ybar=9pt,
	bar width=12pt,
	symbolic x coords={FOOD,AREA,PRICE,all},
	nodes near coords,
]
\addplot 
	coordinates {(FOOD,90.6) (AREA,96.7)
		 (PRICE,97.7) (all,85.9)};

\addplot 
	coordinates {(FOOD,91.1) (AREA,97.2)
		 (PRICE,97.9) (all,87.0)};

\legend{TEN--XH,TEN--X}
\end{axis}
\end{tikzpicture}
			}
		\end{tabular}
	\end{center}
	\caption{turn-level state accuracy for TEN--XH and TEN--X on the DSTC2 dataset. The {\em PRICE} indicates the ${\tt PRICERANGE}$ slot. The {\em all} denotes the proportion
of dialogue turns that the turn-level states for all slots are correctly predicted.}
	\label{fig:bar}
\end{figure}
\subsection{Temporal Analysis}
To analyze how the temporal dependencies influence the state tracking performance, we report the joint goal accuracy at each dialogue turn on the DSTC2 dataset. As shown in Figure~\ref{fig:seq}, the joint goal accuracy of proposed models generally decrease at earlier turns and increase at later turns, as the turns increase. This phenomenon can be explained by the fact: in the earlier stage of the dialogue, more slots are involved in the conversation as the dialogue progress; thus more slot-value pairs need to be predicted in state estimation, making the state harder to calculate correctly; in the later stage of dialogue, the state becomes fixed because the values for all slots are already determined, making the state easier to predict. 
Another observation is that the gaps between TEN--XH and TEN generally increase as the turns increase, showing that modeling temporal dependencies reduces state estimation errors, especially when the dialogue is long. By modeling temporal feature dependencies and temporal state dependencies respectively, TEN--Y and TEN--X also perform better than TEN--XH as the turns increase.
\subsection{Effectiveness of the Hierarchical Encoder}
To prove the effectiveness of the Hierarchical Encoder module, we report the turn-level state accuracy for TEN--XH and TEN--X on the DSTC2 dataset. From the results in Figure~\ref{fig:bar}, we observe that TEN--X, with the Hierarchical Encoder module, achieves higher turn-level state accuracy than TEN--XH for all slots. 

Recall that TEN--X achieves higher joint goal accuracy than TEN--XH, we could think that the performance gain for TEN--X is due to its improvement in turn-level state prediction. This fact demonstrates the significance of considering temporal feature dependencies in turn-level state prediction and illustrates the effectiveness of the Hierarchical Encoder module in TEN--X.

\subsection{Effectiveness of the Belief Propagation}
Table~\ref{tab:ex_bp} is an example of dialogue state tracking selected from the test set of the DSTC2 dataset. As we observe from the table, at turn 1 and turn 2, the user does not specify any food type; both TEN--X and TEN correctly predict the true value ${\tt unknown}$. At turn 3, the user expresses that he or she does not care about the food type. This time the turn-level state predictor gets an incorrect turn-level state value ${\tt unknown}$, instead of the correct one ${\tt dontcare}$. Thus TEN--X gets a wrongly aggregated state value ${\tt unknown}$ with aggregating rule $\triangleleft$. On the contrary, TEN can still correctly obtain the correct state with the belief propagation, in spite of the wrong turn-level state. At turn 4, the turn-level state predictor easily predicts the correct value ${\tt unknown}$ and TEN keeps the state correct. But TEN--X fails to obtain the correct state again because of the wrong decision made at the last turn. This example shows the effectiveness and robustness of the state aggregation approach equipped with the belief propagation.

\section{Related Works}
\label{sec:related}
Traditional works deal with the DST task using Spoken Language Understanding (SLU), including~\cite{Thomson:10,Wang:13,Lee:16,Liu:17a,Jang:16,Shi:16,Vodol:17}. 
Joint modeling of SLU and DST~\cite{Henderson:14,Zilka:15,Mrksic:15} has also been presented and shown to outperform the separate SLU models. 
Models like~\cite{Sun:14,Yu:15} incorporate statistical semantic parser for modeling the dialogue context. These models rely on hand-crafted features or delexicalisation strategies and are difficult to scale to realistic applications. 

Recently, neural network models have been applied in the DST task, and there are mainly two model design approaches. One approach aggregates the features extracted from previous turns of the dialogue using recurrent neural networks, including StateNet~\cite{Ren:18}, LargeScale\cite{Ramadan:18} and SUMBT~\cite{Lee:19}. The other approach, like NBT~\cite{Mrksic:17} and GLAD~\cite{Zhong:18}, build a model for predicting turn-level state, and estimate the state by accumulating all previous turn-level states. The design of TEN integrates the advantages of both approaches.


Another topic related to our work is the Markov decision process (MDP) and the factor graphs. Several works define a dialogue system as a partially observable Markov decision process (POMDP), including~\cite{Williams:07,Thomson:10,Gasic:11,Yu:15}. In this paper, the definition of the dialogue process is related to the Markov decision process. The factor graphs have been applied in many applications, such as social influence analysis~\cite{Tang:09}, knowledge base alignment~\cite{Wang:12}, entity linking~\cite{Ran:18} and visual dialog generation~\cite{Schwartz:19}. The factor graphs in these applications are used to integrate different sources of features or representations into a unified probabilistic model. In this paper, the factor graphs are naturally adopted to tackle the error propagation problem in state aggregation.
\section{Concluding Remarks}
Our inspiration for TEN comes from a careful study of the dialogue process. This allows us to lay out the dependency structure of the network as in Figure~\ref{fig:mod} (c), where the temporal feature dependencies and the temporal state dependencies are jointly modelled. The application of the belief propagation in this model allows an elegant combination of graphical models with deep neural networks. The proposed model may generalize to other sequence prediction tasks.

\section*{Acknowledgment}
This work is supported partly by the National Natural Science Foundation of China (No. 61772059, 61421003), by the Beijing Advanced Innovation Center for Big Data and Brain Computing (BDBC), by the Fundamental Research Funds for the Central Universities and by the Beijing S\&T Committee (No. Z191100008619007) and by the State Key Laboratory of Software Development Environment (No. SKLSDE-2020ZX-14).
\bibliographystyle{acl_natbib}
\bibliography{emnlp2020}

\end{document}